\documentclass[sigconf]{acmart}

\AtBeginDocument{%
  }

\usepackage{multirow}
\usepackage{makecell}

\begin{document}

\title{Evaluating Knowledge Generation and Self-Refinement Strategies for LLM-based Column Type Annotation}

\author{Keti Korini}
\orcid{0000-0002-2158-0070}
\affiliation{%
  \institution{Data and Web Science Group\\University of Mannheim}
  \city{Mannheim}
  \country{Germany}
}
\email{kkorini@uni-mannheim.de}

\author{Christian Bizer}
\orcid{0000-0003-2367-0237}
\affiliation{%
  \institution{Data and Web Science Group\\University of Mannheim}
  \city{Mannheim}
  \country{Germany}}
\email{christian.bizer@uni-mannheim.de}

\renewcommand{\shortauthors}{Korini and Bizer}

\begin{abstract}
Understanding the semantics of columns in relational tables is an important pre-processing
step for indexing data lakes in order to provide rich data search. An approach to establishing such understanding is column type annotation (CTA) where the goal is to annotate table columns with terms from a given vocabulary. This paper experimentally compares different knowledge generation and self-refinement strategies for LLM-based column type annotation. The strategies include using LLMs to generate term definitions, error-based refinement of term definitions, self-correction, and fine-tuning using examples and term definitions. We evaluate these strategies along two dimensions: effectiveness measured as F1 performance and efficiency measured in terms of token usage and cost. 
Our experiments show that the best performing strategy depends on the model/dataset combination. We find that using training data to generate label definitions outperforms using the same data as demonstrations for in-context learning for two out of three datasets using OpenAI models.
The experiments further show that using the LLMs to refine label definitions brings an average increase of 3.9\% F1 in 10 out of 12 setups compared to the performance of the non-refined definitions. 
Combining fine-tuned models with self-refined term definitions results in the overall highest performance, outperforming zero-shot prompting fine-tuned models by at least 3\% in F1 score.
The costs analysis shows that while reaching similar F1 score, self-refinement via prompting is more cost efficient for use cases requiring smaller amounts of tables to be annotated while fine-tuning is more efficient for large amounts of tables.

\end{abstract}

\keywords{Table Annotation, Column Type Annotation, Large Language Models, Self-refinement, Self-correction, Fine-tuning}
\maketitle

\section{Introduction}

\begin{figure}
  \centering
  \includegraphics[width=\linewidth]{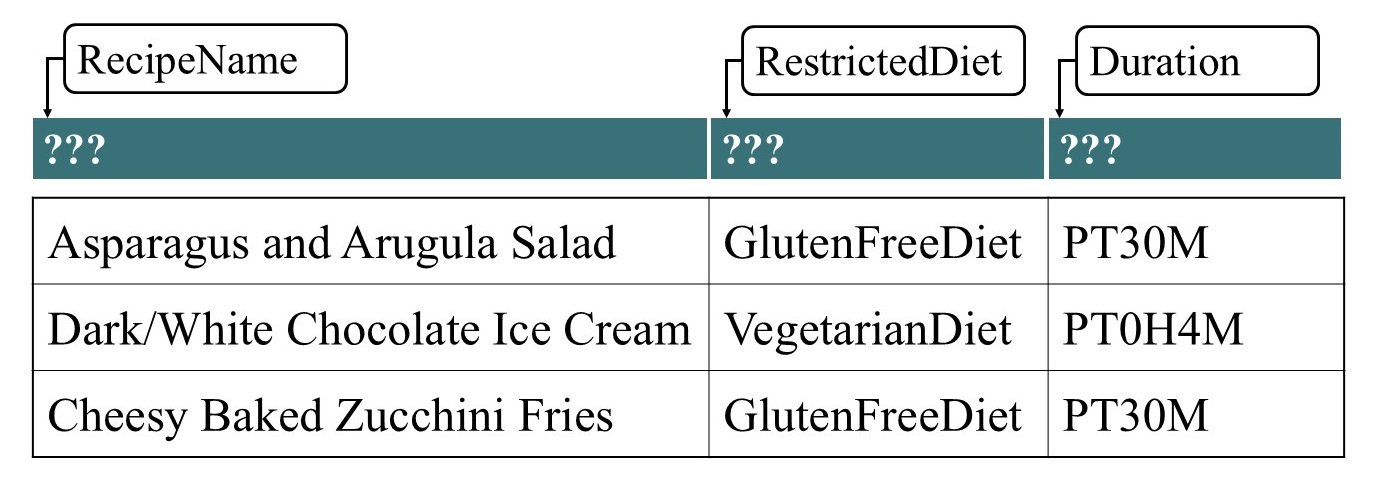}
  \caption{An example of a table that is annotated with column types. The CTA labels are shown above the columns.}
  \label{fig:cta}
\end{figure}

With the growing number of datasets that are collected within corporate data lakes, data search and discovery methods are increasingly becoming critical to enable organizations to take advantage of the collected data~\cite{chapman2020dataset,ChristensenLLRM25}.
However, identifying relevant data faces challenges such as schema heterogeneity~\cite{nargesian2019data,limaye2010annotating} and lack of table metadata~\cite{chapman2020dataset}. Annotating tabular data with terms from a single vocabulary can help alleviate these challenges~\cite{venetis2011recovering, khatiwada2023santos}. 
Tasks that target the understanding of table semantics are grouped under the concept of \textit{table interpretation}~\cite{taSurvey2023}. 
Column Type Annotation (CTA) 
is a sub-task of table interpretation where the goal is to link columns in a table to semantic types from a pre-defined vocabulary based on the type of the entities contained in the column. An example of CTA is shown in Figure \ref{fig:cta}. In the table, the first column can be linked to the ``RecipeName" label as it contains entities that describe the names of different recipes.

To tackle the task of CTA, different works build deep learning frameworks~\cite{sherlock-2019,sato-2020,chen2019colnet,chen2019learning}, leverage knowledge graphs (KGs)~\cite{jimenez-ruiz_semtab_2020,nguyen2019mtab,liu2019dagobah},
or fine-tune pre-trained language models (PLMs)~\cite{trabelsi2021selab,deng_turl_2022,suhara2022annotating}, such as BERT~\cite{devlinBERTPretrainingDeep2019}. With the recent advancements in the field of NLP, large language models (LLMs) have been used in different setups on the task of CTA and have shown great promise in dealing with the need for large amounts of labeled data that the previous PLMs required. In these works, prompting techniques are explored~\cite{KoriniCtaVldb2023,kayali2023chorus}, as well as fine-tuning LLMs either specifically for the task of CTA or for several table tasks including CTA~\cite{li2023table,zhang2024tablellama,feuer2023archetype}.

In tasks such as code generation and commonsense reasoning, LLMs have shown the ability to  \textit{self-refine}~\cite{madaan2023self} and \textit{self-correct}~\cite{pan2023automatically} their output, meaning that the model improves its initial answer based on external feedback or feedback generated by the model itself. While self-refinement and self-correction  are reported to improve results in some experiments,~\cite{olausson2023self,huang2024large} report that the techniques increase processing costs without improving results. \cite{brinkmann2025automated} has encountered the same problem when applying self-refinement and self-correction for information extraction.

In this paper, we investigate the utility of knowledge generation prompting, self-refinement, and self-correction methods for the table column annotation task (CTA). We compare their effectiveness in terms of F1 score and efficiency in terms of token usage and costs to alternative approaches such as self-consistency prompting~\cite{wang2023selfconsistency} and fine-tuning. 
We experiment with knowledge prompting by generating label definitions for the vocabulary terms of the datasets in order to adapt the knowledge of the LLMs to the specific meaning of the labels in the datasets.
In order to test the \textit{self-refinement} abilities of the LLMs, we explore an error-based refinement method where we refine the previously generated definitions using as signals errors made during the annotation of the validation set.
To test the \textit{self-correction} abilities of LLMs, we experiment with a two-step pipeline, where in the first step an LLM annotates the columns of the test set, while in the second step we initialize an LLM with the role of a reviewer to review and correct the annotations of the first model.

While previous papers~\cite{li2023table,zhang2023jellyfish} focused on training generic models to enhance performance across multiple table-related tasks, we focus on fine-tuning LLMs specifically for the CTA task.
As in real-world use cases labeled data is expensive to obtain and table metadata can be missing, we focus on a scenario with low amounts of training data and no table metadata, such as headers or table captions. 
 For this scenario, we choose three datasets which cover different topical domains and both multi-class and multi-label CTA: The Limaye~\cite{limaye2010annotating}, WikiTURL~\cite{deng_turl_2022}, and SOTAB V2~\cite{korini_sotab_2022} datasets. We subsample and restrict the number of examples per label for the last two datasets, as the original training sets include more than 40,000 tables.
We evaluate our methods using four large language models (LLMs): two open-source models \textit{Llama-3.1-8B-4-bit} and \textit{Llama-3.1-70B-4-bit}, and two closed-source models \textit{gpt-4o-mini-2024-07-18} and \textit{gpt-4o-2024-03-15}. 
The main findings of the paper are:
\begin{itemize}
   \item \textbf{There is no single best strategy}: Our experiments show that there is no single best knowledge generation or self-refinement strategy, but the best performing method depends on the model/dataset combination. This finding is inline with findings from related work which come to the same conclusion concerning the usage of LLMs for the entity matching task~\cite{peetersEntityMatching2025}.
   \item \textbf{Knowledge generation prompting improves the performance of OpenAI models compared to five-shot prompting}: Generating label definitions using demonstrations from the training set improves the performance of the CTA task 0.8-9.5\% F1 on the OpenAI models for 2 out of 3 datasets compared to using the same demonstrations for five-shot in-context learning. This increase in F1 score, comes with an increase of 1.3-6 times the \textit{inference} tokens of five-shot prompting.
   
    \item \textbf{Self-refinement brings an increase to the performance of most models tested}: We explore refining the generated label definitions by using errors from the validation set. This increases the F1 score in 10 out of 12 cases by at least 1.5\% and an average of 3.9\% compared to non-refined definitions. However, the additional refinement step increases the \textit{generation} token usage by more than 40 times the token usage of the non-refining scenario.
    \item \textbf{Self-correction does not improve the performance in most cases}: Testing our two-step self-correction pipeline on three different datasets using four models, we find that the model performance is harmed compared to simple one-step CTA in more than half the total 36 cases that we test. Pairing the pipeline with label definitions, we find that for two datasets out of three, when using the larger \textit{gpt-4o} model, we have an increase in F1 score by at least 1.7\%.

    \item \textbf{Fine-tuning LLMs for CTA is more token efficient for large use cases}: 
    For a comparison to the self-refinement pipeline, we experiment with fine-tuning four LLMs on the task of CTA. Fine-tuning brings an increase of at least 6.3\% in F1 score in 7 out of 8 cases tested.
    While we find out that we have a high token usage when fine-tuning the models, the tokens used in inference are 6-10 times less for the fine-tuning scenario than that of the self-refinement scenario with no fine-tuning. This means that for use cases requiring a large amount of tables to be annotated, fine-tuning is more efficient than self-refinement via prompting.
    \item \textbf{Combining fine-tuning and self-refinement is the best performing method for gpt-4o}: By generating and refining label definitions with the fine-tuned models, we find that using the fine-tuned \textit{gpt-4o} model in combination with the refined definitions via prompting brings an additional
    3\% increase when compared to using the fine-tuned model with zero-shot prompting in the two datasets that we use in our fine-tuning experiments. However, this improvement comes at the cost of a 4- to 8-fold increase in token usage. For the smaller models tested, the self-refinement is not efficient both in tokens and in performance.
\end{itemize}

The paper is structured as follows: In Section~\ref{sec:rel-work}, we discuss related work, followed by the description of our experimental setup in Section~\ref{sec:exp-setup}. In Section~\ref{sec:baselines}, we test zero-shot, few-shot and self-consistency prompting to establish baselines. In Section~\ref{sec:labels}, we test knowledge prompting by generating three types of label definitions and how to use LLMs to self-refine definitions. In Section~\ref{sec:correction}, we test a two-step self-correction pipeline for the task of CTA. In Section \ref{sec:ft}, we evaluate different fine-tuning setups including knowledge generation. Section~\ref{sec:conclusion} summarizes our conclusions.

\section{Related Work}
\label{sec:rel-work}

We organize the related work section in four parts: table annotation systems, column type annotation methods, LLM-based column type annotation, and general prompting techniques.

\textbf{Table Annotation.} 
Numerous works~\cite{taSurvey2023, BalogSurvey2000,wu2016entity,efthymiou2017matching,venetis2011recovering} have been published that tackle different sub-tasks of table annotation such as CTA, column property annotation (CPA) and cell entity annotation (CEA). A line of these works~\cite{liu2019dagobah,nguyen2019mtab,cremaschi2022s,sarma2012finding,jiomekong2022towards,avogadro2021mantistable} leverage knowledge bases (KBs) to discover semantics of the tables by firstly linking entities in a table to entities found in KBs and thus determining the column type and column relationships by looking at the type and properties of the entity in the KB. The Semantic Web Challenge on Tabular Data to Knowledge Graph Matching (SemTab)~\cite{jimenez-ruiz_semtab_2020} is a challenge that was designed to yearly benchmark table annotation systems, like DAGOBAH~\cite{liu2019dagobah} and MTab~\cite{nguyen2019mtab}, on different datasets. Other works rely on building table representation models and use either cell or column embeddings to classify the column types or relationships. In this category, in TURL~\cite{deng_turl_2022} a TinyBERT~\cite{jiao-etal-2020-tinybert} model is further pre-trained on table corpora and learns cell representations which are then used on the tasks of CTA, CPA and CEA. TABBIE~\cite{iida-etal-2021-tabbie} uses two Transformers~\cite{vaswani-transformer-2017} to learn row and column representations independently by further pre-training a BERT~\cite{devlinBERTPretrainingDeep2019} model with a new pre-training objective: the Cell Corruption Detection objective.

\textbf{Column Type Annotation.} Early CTA systems that use deep learning frameworks, like Sherlock~\cite{sherlock-2019} and Sato~\cite{sato-2020}, use feature engineering to construct features from statistical and character-level column values information. With the evolution of the Transformer~\cite{vaswani-transformer-2017} architecture, PLMs were mainly used for CTA. SeLab~\cite{trabelsi2021selab} fine-tunes a BERT~\cite{devlinBERTPretrainingDeep2019} model, treating a column as a sequence to feed to the PLM. 
In Doduo~\cite{suhara2022annotating} a BERT model is fined-tuned using multi-task learning for CTA and CPA and contributes the idea of serializing a whole table. TorchicTab~\cite{Dasoulas2023TorchicTabST}, uses Doduo as a base architecture and extends it by making it possible to include non-labeled columns as context and handle wide tables. Recently, there has been focus on CTA systems that improve the performance on numerical columns, such as Pythagoras~\cite{langenecker2024pythagoras} and SAND~\cite{su2023sand}.

\textbf{LLM-based methods for CTA.} 
In \cite{KoriniCtaVldb2023} prompt designs that include instructions and different formulations of the CTA task are explored using ChatGPT~\cite{honovich2022instruction}. Chorus~\cite{kayali2023chorus} explores three table tasks including CTA and the idea of “anchoring” predictions is presented, where out-of-vocabulary answers are embedded and using the nearest neighbor method, are mapped to the closest labels in the label space.
In ArcheType~\cite{feuer2023archetype} zero-shot CTA is explored alongside fine-tuning a Llama-7B model on the task of CTA. 
The authors evaluate different ways of sampling rows, serializing tables and like Chorus propose their own method of remapping the out-of-vocabulary answers of the model. In RACOON~\cite{wei2024racoon} a retrieval augmented method is explored where prompts are augmented with relevant information retrieved from a KG. 
In Table-GPT~\cite{li2023table}, the authors aim at creating a generalist model in order to increase the performance across multiple table tasks. For this they fine-tune ChatGPT on general structure understanding, such as retrieving certain rows or columns and on a variety of table tasks such as schema matching, entity matching and data imputation. Similarly, in TableLlama~\cite{zhang2024tablellama} a Llama-7B model is fine-tuned on six table tasks, two of which are CTA and CPA. In Jellyfish~\cite{zhang2023jellyfish} a Llama-2-13B variant further pre-trained on the Open-Platypus dataset is fine-tuned on four tasks such: error detection, data imputation, schema matching and entity matching and the model is tested on CTA. 
In contrast to existing papers on prompting and fine-tuning LLMs for the task of CTA, in our paper we focus on investigating the use of knowledge generation prompting and self-refinement strategies and how we can use them in both a non-fine-tuning and fine-tuning scenario. We focus not on generalization, but on improving the performance of the task of CTA. A following work to Table-GPT, Table-specialist~\cite{xing2024table} notes that generalist models come at the cost of performance loss for individual table tasks and build specialist models. However, CTA is not experimented on.

\textbf{Prompting techniques.} Knowledge generation prompting is explored by Liu et al.~\cite{liu2021generated} to generate knowledge facts about commonsense reasoning tasks. They conduct knowledge generation in two steps, where in the first step a T5 and a GPT-3 model are prompted to output some knowledge about questions while being presented with some demonstrations and in the second step they augment their prompts with this generated knowledge. 
For tasks such as code generation and reasoning, the concept of \textit{self-correction}~\cite{pan2023automatically} is tested with the goal of correcting in a post-processing step the initial output of the LLMs. In~\cite{madaan2023self}, tasks such as sentiment reversal, dialogue response, code optimization and readability benefit from the self-refinement as the LLMs successfully refine previous answers, while the task of math reasoning is not improved. Paul et al.~\cite{paul2024refiner} introduce their system REFINER where intermediate reasoning steps are generated for mathematical questions and a critic model provides feedback on this reasoning leading to an increased performance on this task. Other following works~\cite{olausson2023self,huang2024large}, however, point that the self-correction ability does not show consistent behavior among datasets. They point that the technique is expensive as the model should be prompted more than once and the improvement in the tasks were due to the non-optimal prompts used in the first run of the model.

\section{Experimental Setup}
\label{sec:exp-setup}
This section introduces the datasets and models that we use in the experiments as well as the evaluation metrics. The code, prompts, data, and responses from the LLMs are available in the Github repository\footnote{\url{https://github.com/wbsg-uni-mannheim/TabAnnGPT}} accompanying this paper.

\begin{table}
  \caption{Datasets' statistics: number of tables in the training, validation and test splits as well as the average number of annotated columns per table (Avg.C), number of labels (Lab.) and the number of annotated columns in the test set (Test.C).}
  \label{tab:datasets}
  \begin{tabular}{lcccccc}
    \toprule
    \textbf{Dataset} & \textbf{Train} & \textbf{Val.} & \textbf{Test} & \textbf{Avg.C} & \textbf{Lab.}  & \textbf{Test.C}\\
    \midrule
    SOTAB V2 & 44K & 456 & 609 & 2.62 & 82  & 1,851\\
    SOTAB V2-ds & 698 & 199 & 239 & 2.80 & 50  & 824 \\
    WikiTURL & 397K & 4,8K & 4,7K & 1.60 & 255  & 13K \\
    WikiTURL-ds & 809 & 416 & 379 & 1.44 & 66  & 878 \\
    Limaye & 105 & - & 107 & 1 & 26 & 107 \\
  \bottomrule
\end{tabular}
\end{table}

\textbf{Datasets.} In our experiments we use three datasets: SOTAB V2 CTA\footnote{\url{https://webdatacommons.org/structureddata/sotab/v2/}}, the WikiTables-TURL~\cite{deng_turl_2022} dataset, which we will refer as WikiTURL throughout the paper, and the Limaye~\cite{limaye2010annotating} dataset. \textit{SOTAB V2 CTA} is a large heterogeneous CTA dataset with tables that were collected from 44,268 different websites and are spread over 17 topical domains. Altogether, the SOTAB V2 
dataset contains over 40,000 training tables. For our experiments, we down-sample the dataset by choosing five domains out of the 17: Book, Recipe, Restaurant, Movie and Product. 
To work in the scenario with limited amounts of training data, 
we reduce the number of training examples in the training set to 
20 columns per label resulting in 698 training tables (see Table \ref{tab:datasets}). 
\textit{WikiTURL CTA} is a large dataset created by Deng et al.~\cite{deng_turl_2022} by annotating tables from
the WikiTables corpus~\cite{bhagavatula2015tabel} using Freebase~\cite{bollacker2008freebase}. Similarly to SOTAB V2, we down-sample WikiTURL by limiting tables to five domains: film, music, book, food and broadcast. 
We again limit the amount of training data to 20 columns per label resulting in 809 training tables. The last dataset we use is the \textit{Limaye} dataset~\cite{limaye2010annotating}. The tables are gathered from Wikipedia pages and we use the version annotated by Efthymiou et al.~\cite{efthymiou2017matching} using the DBpedia~\cite{auer2007dbpedia} vocabulary. 
We split the 212 total tables into a training and test set, and use the training set as a validation set when necessary. The main statistics of all datasets are summarized in Table \ref{tab:datasets}. For SOTAB V2 and WikiTURL, we show the statistics for both the full and down-sampled sets, marked with the suffix \textit{-ds} in the table.

\textbf{Models.} To test our methods, we select four models: two open-source Llama-3.1 models, \textit{Llama-3.1-8B-4-bit}\footnote{\url{https://huggingface.co/unsloth/Meta-Llama-3.1-8B-Instruct-bnb-4bit}} and \textit{Llama-3.1-70B-4-bit}\footnote{\url{https://huggingface.co/unsloth/Meta-Llama-3.1-70B-Instruct-bnb-4bit}} and two OpenAI models, \textit{gpt-4o-mini-2024-07-18} and \textit{gpt-4o-2024-03-15}. For the fine-tuning experiments with gpt-4o we use the available \textit{gpt-4o-2024-08-06} model. For brevity, in the paper we will refer to these models as Llama-8B, Llama-70B, gpt-mini and gpt-4o respectively. For fine-tuning the open-source models we use the Unsloth\footnote{\url{https://docs.unsloth.ai/}} and Huggingface\footnote{\url{https://github.com/huggingface/transformers}} libraries while for the OpenAI models we use the Langchain\footnote{\url{https://www.langchain.com/}} library. We set the temperature of the models to 0 for OpenAI models and 0.001 for the Llama-3.1 models in order to reduce randomness. 

\textbf{Serialization.} The tables in the prompts are serialized in markdown format. We include only the first five rows of a table in the prompts. We  restrict the number of words in the table cells to 20 words. In datasets such as Limaye and SOTAB V2 where context columns (columns not annotated) are available, we include the columns as well and we ask for the classification only of the columns in the test set. In WikiTURL and Limaye, where column headers as well as table description text is available, we do not use this information as we want to focus on a scenario with missing metadata. We therefore substitute the column names to ```Column 1", ``Column 2", etc.

\textbf{Evaluation Metrics.} The CTA task in the SOTAB V2 dataset is treated as a multi-class classification problem with a class imbalance. For this reason, we use as our main metric the Micro-F1 score. The WikiTables-TURL and Limaye datasets treat CTA as a multi-label classification problem and also include class imbalance. Therefore, in addition to Micro-F1 we also report the Hamming score as an additional multi-label metric. Both open-source models are run 3 times and we report the average metrics of the three runs. Similarly, for our fine-tuning experiments, we run fine-tuning with open-source models three times with three different seeds and report the average metric. We do not re-map the out-of-vocabulary answers to the label space but treat the out-of-vocabulary answers as errors. In addition to F1, we report precision and recall for all experiments in the GitHub repository.

\section{Zero-shot, Few-shot and Self-consistency Baselines}
\label{sec:baselines}

\begin{figure}
  \centering
  \includegraphics[width=\linewidth]{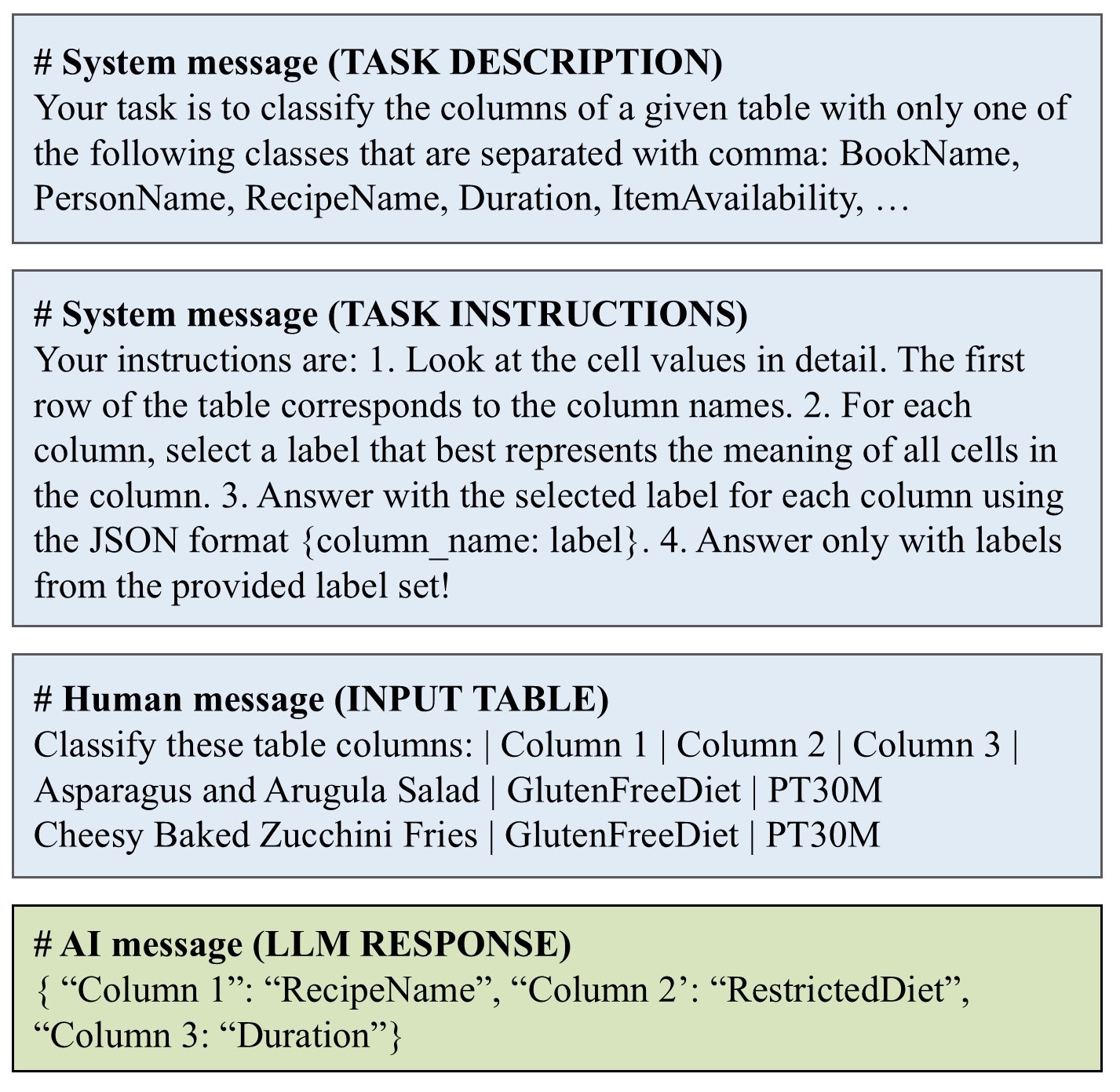}
  \caption{Example of a zero-shot prompt.}
  \label{fig:zero-shot}
\end{figure}

In this section, we present our baselines and the results on the chosen datasets. 
Our first baseline is \textit{zero-shot} prompting. A zero-shot prompt includes a \textit{task description} where the label set is presented, \textit{task instructions} on how to carry the annotation and the JSON format the model should respond with, and the \textit{input table}. In the case of Llama models, we do not include task instructions in the prompt as initial experiments showed that instructions harmed the performance of these models. An example of a zero-shot prompt is shown in Figure~\ref{fig:zero-shot} which includes the serialization of the table in Figure~\ref{fig:cta} and an example response from the LLM (\textit{assistant message}). 
The second baseline is \textit{few-shot} prompting, where we present to the model 5 demonstrations picked from the training set. We pick the demonstrations for each test table based on the similarity between test and training tables. To compute this similarity, we embed the test and training tables by using OpenAI's \textit{text-embedding-3-small} model, without including the ground truth labels. We then compute the cosine similarity between the embedding vectors and choose for each test table the 5 most similar training tables to use as demonstrations to the LLM. Our third baseline is \textit{self-consistency} prompting introduced by Wang et al.~\cite{wang2023selfconsistency}. This technique consists of prompting the model multiple times to generate different answers and use majority voting for selecting the final answer. In our experiments, we run the LLMs 3 times with three different temperatures, 0, 0.5 and 0.7, and for each column we select the label that was predicted more than once in these three runs.

\begin{table}
  \caption{Baselines' Micro-F1 results on three datasets.}
  \label{tab:baselines}
  \begin{tabular}{llcccc}
    \toprule
    \textbf{Dataset} & \textbf{Setup} & \textbf{Ll-8B} & \textbf{Ll-70B} & \textbf{gpt-mini} & \textbf{gpt-4o}  \\
    \midrule
    \multirow{3}{*}{SOTAB V2}  & 0-shot & 56.0 & 67.4 & 69.4 & 80.9 \\
    & self-con. & 55.2 & 67.5 & 70.6 & 80.0 \\
     & 5-shot & 59.3 & 64.6 & 63.4 & 81.8 \\
     \midrule
    \multirow{3}{*}{Limaye} & 0-shot & 66.0 & 76.7 & 76.5 & 82.6 \\
    & self-con. & 66.7 & 77.8 & 80.8 & 84.3 \\
     & 5-shot & 90.2 & 81.5 & 91.2 & 89.4 \\\midrule
     \multirow{5}{*}{WikiTURL} & 0-shot & 15.0 & 54.0 & 53.3 & 55.3 \\
     & 5-shot & 40.4 &  54.8 & 56.0 & 61.5 \\
     & 0-hier & 31.8 & 57.3 & 61.0 & 70.0 \\
     & self-con. & 36.3 & 58.4 & 61.6 & 69.2 \\
     & 5-hier & 43.6 & 56.1 & 58.5 & 63.7 \\
  \bottomrule
\end{tabular}
\end{table}

\textbf{Results.} The results of the baseline experiments are shown in Table~\ref{tab:baselines}. Observing the WikiTURL label set, we notice that the labels follow a hierarchy, therefore we manually construct a label hierarchy without using the ground truth, and experiment by including this hierarchy information in the prompts. In the table, we denote with \textit{-hier} the results of adding the hierarchy information in the prompt. We observe that by including the hierarchy information for all models we have a large increases from 3.3-16.8\% in F1 score in the zero-shot setting, with the larger increases (>10\% F1) for Llama-8B and gpt-4o. Although the increase becomes smaller in the few-shot prompting case, the models benefit with an additional 1.3-3.2\% F1.

A general pattern that we observe is that in SOTAB V2 and WikiTURL, including 5 demonstrations improves the performance of the smaller Llama-8B model over the zero-shot scenario. On the other hand, when these demonstrations are fed to larger models, the F1 score is either similar or less than the zero-shot scenario.
In contrast, in the Limaye dataset in all cases when we include 5-shots the F1 increases to around 90\%. To understand the difference in this behavior, we calculate the average cosine similarity of the demonstrations we include in the prompts for all datasets. We find out that the average similarity of the test tables to the 5 selected demonstrations is higher in Limaye, with an average cosine similarity of 0.755 than in SOTAB V2 and WikiTURL where this average similarity is 0.654 and 0.648 respectively. This smaller similarity could be the reason of the smaller increases of few-shot prompting for the SOTAB V2 and WikiTURL datasets. Regarding self-consistency prompting, we notice that in most cases we have similar results to zero-shot prompting.

\textbf{Error Analysis.} To gain more insight into the differences between the methods, we conduct an error analysis of the results of zero-shot prompting and five-shot prompting scenarios on the SOTAB V2 dataset. The labels with more than 5 errors are shown in Table~\ref{tab:errors-zero} for both methods. We notice the errors have a long-tail distribution, where 90\% of the labels have 5 or less errors. In zero-shot prompting, there are 10 labels out of 50 where \textit{gpt-4o} has made more than 5 errors, while in few-shot prompting this number falls to 6 labels. Comparing both methods, we notice that 7 labels out of 10 are improved with five-shot prompting. The increase however of the errors for the \textit{Photograph} label (from 1 to 14 errors) in the five-shot scenario makes the final difference between both methods only 0.9\% F1. We find out that the \textit{Photograph} label which is used to annotate columns that include URLs pointing to images, are confused with the label \textit{URL}. The demonstrations in the prompts did not help the LLM as the embedding based search for similar columns picked \textit{URL} columns instead of \textit{Photograph} columns to show as demonstrations. We report the full set of errors in our repository. 

\begin{table}
  \caption{Number of inference tokens (Inference) and costs (Infer. Cost), and annotation cost per column (Cost/Column) of baselines for gpt-4o.}
  \label{tab:cost-zero-few-self}
  \begin{tabular}{lrrrr}
    \toprule
     \textbf{Dataset} & & \textbf{0-shot} & \textbf{self-cons.} & \textbf{5-shot}\\
    \midrule
    \multirow{2}{*}{\textbf{SOTAB V2}} & Inference & 270K & 810K & 1,047K \\
    & Infer. Cost & \$0.87 & \$2.61 & \$2.73\\
    & Cost/Column & \$0.001 & \$0.004 & \$0.004 \\
    \midrule
    \multirow{2}{*}{\textbf{Limaye}} & Inference & 33K & 99K & 90K\\
    & Infer. Cost & \$0.09 & \$0.28 & \$0.23 \\
    & Cost/Column & \$0.0008 & \$0.002 & \$0.002 \\
    \midrule
    \multirow{2}{*}{\textbf{WikiTURL}}  & Inference & 476K & 1,428K & 625K\\
    & Infer. Cost & \$1.29 & \$3.89 & \$1.63\\
    & Cost/Column & \$0.001 & \$0.003 & \$0.002 \\
  \bottomrule
\end{tabular}
\end{table}

\textbf{Costs.} We present the costs of the three prompting techniques in Table~\ref{tab:cost-zero-few-self}. We report the total number of (input) tokens required for annotating all tables in the test set, the total inference cost for \textit{gpt-4o-2024-08-06} where 
the input tokens without prompt caching are priced at \$2.5/1M tokens~\footnote{\url{https://openai.com/api/pricing/}} (as of January 2025) and the inference cost per single annotated column (Costs/Column). From the perspective of the F1 score, self-consistency and zero-shot prompting have small differences between them, but on the token perspective, as self-consistency consists of running the LLMs 3 times, the inference tokens are also 3 times higher, making this technique not efficient in both cost and F1 score. Similarly, for 5-shot prompting we have small increases/decreases in F1 score in two out of three datasets for the larger models while the token usage increases by at least 1.3 times.

\begin{table}
    \centering
    \caption{SOTAB V2 labels with more than 5 errors for 0 and 5-shot prompting for gpt-4o, and their difference ($\Delta$0-shot).}
    \begin{tabular}{lcccc}
    \toprule
    \textbf{Label} & \textbf{0-shot} & \textbf{5-shot} & \textbf{$\Delta$0-shot} \\\midrule
         Mass & 25 &  3 &  -22 \\
         Distance & 20 & 9 & -11  \\
         Number & 13 &  9 &  -4 \\
         weight & 11 &  12 & +1 \\
         ItemList & 10 &  1 &  -9 \\
         Time & 10 &  5 & -5  \\
         Energy & 9 &  5 & -4  \\
         ProductModel & 7  & 7 & 0 \\
         uniText & 7 &  9 & +2 \\
         QuantitativeValue & 6 & 2 & -4 \\
         Photograph & 1 &  14 & +13 \\
         \bottomrule
    \end{tabular}
    \label{tab:errors-zero}
\end{table}

\section{Self-Refinement via Knowledge Generation Prompting}
\label{sec:labels}

Liu et al.~\cite{liu2021generated} introduce the concept of \textit{generated knowledge prompting}, where knowledge facts are generated for commonsense tasks to aid in the generation of the answer by the LLM. Inspired by this work, in this section we experiment with the idea of augmenting our prompts with generated label definitions. A label definition is one or more sentences that describe the meaning of a label in the dataset vocabulary and what values it can be used to annotate. The labels can have different, more specific meanings in different datasets influenced by the annotators' guidelines used to create the dataset. These specific meanings can be unknown to the LLMs. Motivated by the idea that we can guide the generation of the annotation based on the specific meaning of a label in a dataset, we test augmenting our prompts with generated label definitions.

\subsection{Definition Generation and Refinement}
\label{sebsec:def-gen}
\begin{figure}
  \centering
  \includegraphics[width=\linewidth]{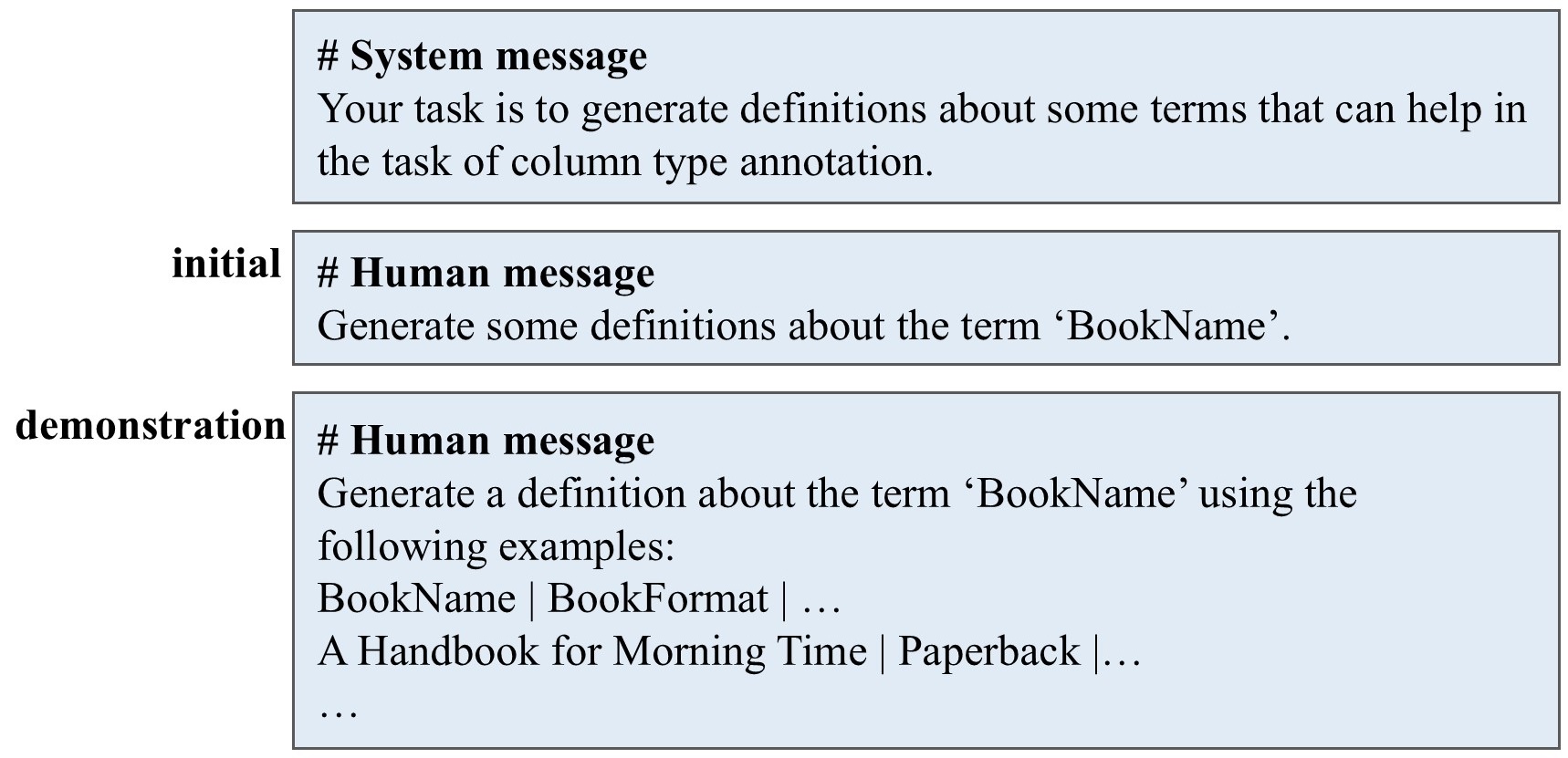}
  \caption{Prompts for definition generation using background knowledge (initial) and demonstrations (demonstration).}
  \label{fig:def-gen}
\end{figure}

We generate three types of label definitions using three different setups which we refer to as \textit{initial} definitions, \textit{demonstration} definitions, and error-based \textit{comparative} definitions. We explain how we generate the three types of definitions below:

\textbf{(1) Initial Definitions.} The initial definitions serve as our baseline definitions and we generate them by prompting the LLM to give a definition for each label in the label set. The initial definitions, therefore, contain the knowledge that the LLM has gained about the terms during its pre-training. An example of the prompt used to generate the initial definition for the label ``BookName" is shown in the first part of Figure~\ref{fig:def-gen}. We feed to the LLM a \textit{system message} containing the task description and a \textit{human message} where we ask the model to generate a definition about the term.

\textbf{(2) Demonstration Definitions.} With the idea to fit the definitions to the meaning that the labels have in a specific dataset, we generate demonstration definitions by using demonstrations from the training set. This means that when prompting the LLM, for each label we show 3 random demonstrations of the label and ask the model to generate the definition based on the shown demonstrations. These definitions reflect, therefore, not only the background knowledge of the LLM about the term, but also the specific usage of the label in the dataset. In Figure~\ref{fig:def-gen}, we show the prompt that we use for generation in the second part of the figure. In this case we feed to the LLM the same \textit{system message} as before that includes the task description, and the \textit{human message} asking the generation of the definition based on the demonstrations shown.

\textbf{(3) Error-based Comparative Definitions.} In this setup, we test the idea of generating comparative definitions that contain tips on how to distinguish different labels from each other. This is motivated by the challenge of labels that have close semantic meanings and which differ in small aspects. 
We generate the comparative definitions by extracting errors that are made when we run CTA on the validation set. This setup consists of three steps: (i) Firstly, we classify the validation set using zero-shot prompting. (ii) Secondly, we extract the columns that were wrongly annotated and group them under their correct label. (iii) Finally, for each label we feed the LLM the false positives and false negatives of the labels, 3 random demonstrations of the label from the training set in order to inform the model of the correct usage of the label, and we ask the model to generate tips on how to distinguish the label from the other labels that it was confused with. For example, if a column is labeled as \textit{RecipeDescription} but its correct label was \textit{Review}, we ask the model to generate tips of how to distinguish between the label \textit{Review} and \textit{RecipeDescription}. We show an example of the prompt for this generation in our repository.

\begin{figure}
  \centering
  \includegraphics[width=\linewidth]{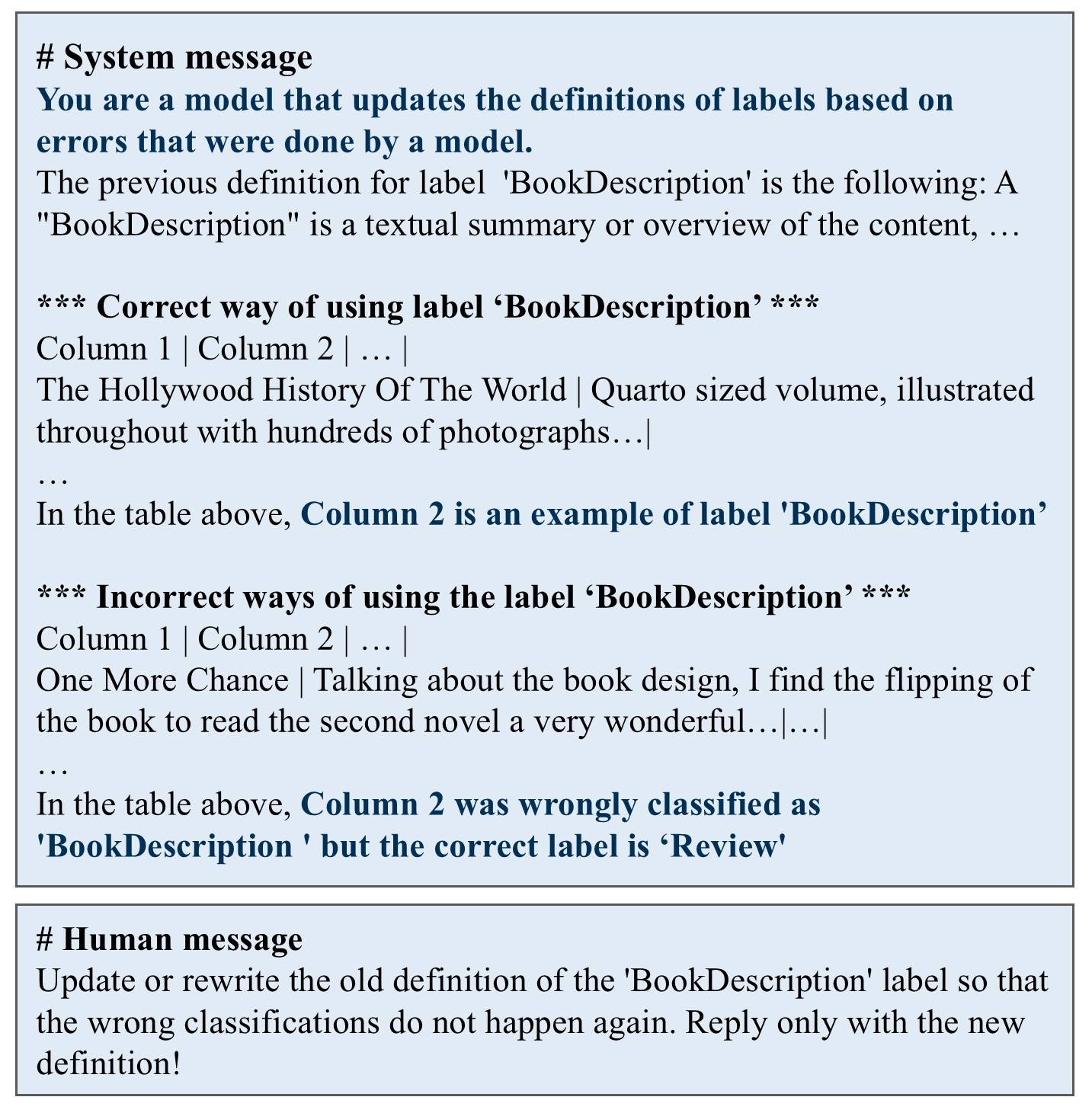}
  \caption{Example of an error-based refinement prompt for the label ``BookDescription".}
  \label{fig:self-ref}
\end{figure}

\textbf{Error-based Refinement.} We further test the idea of the model self-refining the \textit{demonstration} definitions based on errors made on the validation set by the usage of these definitions. This setup consists of 3 steps: (i) Firstly, we classify the validation set by using prompts augmented with the \textit{demonstration definitions}. (ii) Secondly, we extract the errors that were made for each label. (iii) Finally, for each label, we show an LLM the false positives and false negatives of the label, 3 random demonstrations from the training set to show the model the correct way of using the label, the demonstration definition for the label, and we ask the model to update the demonstration definition based on the information given (shown in Figure~\ref{fig:self-ref}). We refer to these definitions as \textit{refined definitions}.

\textbf{Generation details.} We use the \textit{gpt-4o-2024-03-15} model to generate and refine definitions.  
To test the definitions, we use our zero-shot prompts and add the label definitions in the \textit{task description} after we present the label set. 
For the OpenAI models we include all the definitions in the prompt, while for the two Llama models we include only the 10 most similar definitions to the test tables. The similarity is calculated by embedding the definitions as well as the test tables without the ground truth, and selecting for each test table 10 definitions that are most similar by cosine similarity. The model used for generating the embeddings is OpenAI's \textit{text-embedding-3-small} model.  The decision is influenced by initial experiments where the inclusion of all definitions in the prompt for the Llama models damaged the performance of CTA.

\begin{figure}
  \centering
  \includegraphics[width=0.9\linewidth]{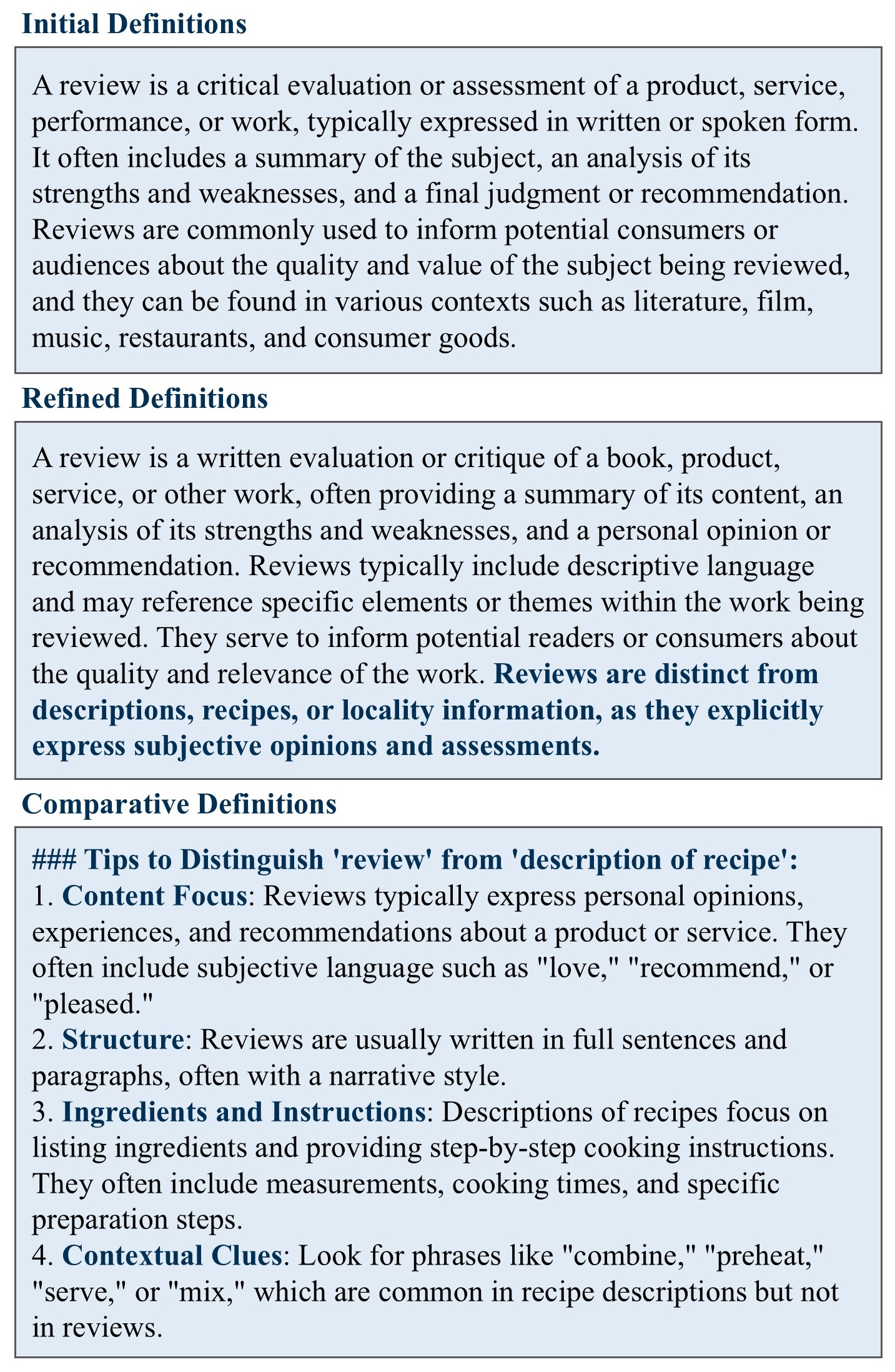}
  \caption{Example of initial, refined and comparative definitions for the label ``Review".}
  \label{fig:def-comp}
\end{figure}

\textbf{Examples of Generated Definitions.} In Figure \ref{fig:def-comp}, we show a textual comparison between \textit{initial}, \textit{refined} and \textit{comparative} definitions for the label ``Review". 
We notice that in both \textit{initial} and \textit{refined} definitions we have a descriptive sentence about the term, while the \textit{comparative} definition is based mostly on keywords and what the column values should contain to be able to be annotated with the label Review. While in \textit{refined} definitions we have mentions of other labels as seen in the last sentence of the definition, the \textit{comparative} definitions include a more detailed analysis of two labels. Finally, the \textit{initial} definition includes a general overview of the meaning of term review, while in the \textit{demonstration} definitions the definition is fitted to the training set as the domains mentioned, books and products, are domains in the SOTAB V2 dataset. We make all the generated definitions available in our repository.

\subsection{Results and Cost Analysis}

\begin{table}
  \caption{SOTAB V2 and Limaye Micro-F1 results testing four types of definitions and the difference of 0 and 5-shot scenario to the best method marked with bold ($\Delta$0-shot/$\Delta$5-shot).}
  \label{tab:definitions}
  \begin{tabular}{llcccc}
    \toprule
    \textbf{Dataset} &\textbf{Setup} & \textbf{Ll-8B} & \textbf{Ll-70B} & \textbf{gpt-mini} & \textbf{gpt-4o}  \\
    \midrule
    \multirow{5}{*}{SOTAB V2} 
    & 0-shot & 56.0 & 67.4 & 69.4 & 80.9 \\
     & initial & 55.0 & 64.3 & 71.7 & 79.3\\
     & demos. & 58.3 & 70.4 & 72.9 & 82.6\\
     & refined & \textbf{59.8} & \textbf{72.0} & \textbf{75.1} & \textbf{85.4} \\
     & compar. & 55.4 & 64.1 & 70.3 & 83.7 \\\midrule
    & $\Delta$0-shot & +3.8 & +4.6 & +5.7 & +4.5 \\
    & $\Delta$5-shot & +0.5 & +7.4 & +11.7 & +3.6 \\
    \midrule
     \multirow{2}{*}{Limaye}
     & 0-shot & 66.0 & 76.7 & 76.5 & 82.6 \\
     & initial & 63.6 & 77.8 & 75.7 & 82.8 \\
     & demos. & 66.1 & 81.6 & 77.7 & 86.8 \\
     & refined & \textbf{79.1} & \textbf{84.9} & \underline{80.2}  & \textbf{88.4} \\
     & compar. & 77.3 & 76.6 & \textbf{80.5} & 85.0 \\\midrule
     & $\Delta$0-shot & +13.1 & +8.2 & +4.0 & +5.8 \\
     & $\Delta$5-shot & -11.1 & +3.4 & -10.7 & -1.0 \\
  \bottomrule
\end{tabular}
\end{table}

\begin{table}
  \caption{Micro F1 (F1) and Hamming Score (HS) results on WikiTURL testing four types of definitions and the difference of 0 and 5-shot scenario to the best method marked with bold ($\Delta$0-sh/$\Delta$5-sh).}
  \label{tab:definitions-turl}
  \begin{tabular}{lcccccccc}
    \toprule
    & \multicolumn{2}{c}{\textbf{Ll-8B}} & \multicolumn{2}{c}{\textbf{Ll-70B}} & \multicolumn{2}{c}{\textbf{gpt-mini}} & \multicolumn{2}{c}{\textbf{gpt-4o}} \\
    \midrule
     \textbf{Setup} & F1 & HS & F1 & HS & F1 & HS & F1 & HS \\\midrule
     0-shot & 31.8 & 25.5 & 57.3 & 52.5 & 61.0 & 51.2 & 70.0 & 63.2 \\
     5-shot & 43.6 & 40.3 & 56.1 & 52.9 & 58.5 & 55.5 & 63.8 & 60.6 \\\midrule
     initial & 29.8 & 23.1 & 47.5 & 44.1 & 60.4 & 50.9 & 69.8 & 65.9  \\
     dem. & 35.3 & 28.8 & \textbf{51.5} & \textbf{47.7} & 61.6 & 52.2 & 72.2 & 67.9 \\
     refine & \textbf{42.6} & \textbf{39.5} & 49.3 & 46.0 & \textbf{65.5} & \textbf{59.9} & \textbf{72.4} & \textbf{68.1} \\
     comp. & 33.2 & 25.6 & 48.8 & 44.9 & 62.2 & 57.1 & 71.6 & 68.7 \\\midrule
     $\Delta$0-sh & +10 & +14 & -5.8 & -4.8 & +4.5 & +8.7 & +2.4 & +4.9 \\
     $\Delta$5-sh & -1.0 & -0.8 & -4.6 & -5.2 & +7.0 & +4.4 & +8.6 & +7.5 \\
  \bottomrule
\end{tabular}
\end{table}

The results of using our four types of definitions on the SOTAB V2 and Limaye datasets are listed in Table \ref{tab:definitions}, while the results for the WikiTURL dataset are listed in Table~\ref{tab:definitions-turl}. From the results, we see that the inclusion of the definitions generated only from the background knowledge of the model harms the performance of all the tested models in three datasets when compared to the zero-shot baseline. On the other hand, including the \textit{demonstration} definitions in the prompts increase the F1 score by an average of 2.4\% in 11 out of 12 setups compared to 0-shot, with the smaller improvement for the Limaye dataset using the Llama-8B model with an increase of 0.1\% and the largest improvement for gpt-4o Limaye results with an increase of 4.2\%. 
The \textit{refined} definitions bring in 10 out of 12 cases an additional average increase of 3.9\% to the F1 score compared to the \textit{demonstration} definitions. Comparing the usage of the \textit{refined} definitions to the zero-shot scenario ($\Delta$0-shot in the tables), we have an overall increase in F1 score of at least 2.4\%. Overall, in 10 out of 12 cases, the \textit{refined} definitions help the models reach the highest F1 score out of all other definitions.
In contrast to our idea that \textit{comparative} definitions can help the models distinguish better between the different labels, \textit{comparative} definitions do not reach a higher F1 score than \textit{refined} definitions in 11 out of 12 cases.

Comparing few-shot prompting to the \textit{demonstration} definitions, we observe that for SOTAB V2 and WikiTURL for the larger models \textit{ gpt-mini} and \textit{gpt-4o}, using demonstrations to generate \textit{demonstration} definitions is more beneficial then using these demonstrations in few-shot prompting, with an increase of 0.8-9.5\% F1. Furthermore, using \textit{refined} definitions brings an increase of at least 3.6\% compared to few-shot prompting for these two models and datasets ($\Delta$5-shot in the tables). For the Limaye dataset, this pattern can not be found as for 3 out of 4 models, the definitions are not better performing than few-shot prompting. 
Overall, we do not find one method that works best for all models and all datasets.

\begin{table}
    \centering
    \caption{SOTAB V2 labels with more than 5 errors for 0-shot and knowledge prompting with refined definitions for gpt-4o, and their difference ($\Delta$0-shot).}
    \begin{tabular}{lcccc}
    \toprule
    \textbf{Label} & \textbf{0-shot} & \textbf{refined} & \textbf{$\Delta$0-shot} \\\midrule
         Mass & 25 &  11 & -14  \\
         Distance & 20 & 17 &  -3  \\
         Number & 13 &  14  & +1 \\
         weight & 11 &  10  & -1  \\
         ItemList & 10 &  1  &  -9 \\
         Time & 10 & 13   &  +3  \\
         Energy & 9 &  8 &  -1  \\
         ProductModel & 7    & 6 & -1 \\
         uniText & 7 &  1  & -6  \\
         QuantitativeValue & 6 &  2 & -4  \\
         \bottomrule
    \end{tabular}
    \label{tab:errors-ref}
\end{table}

\textbf{Error Analysis.} To check in more detail how the error-based refinement method improves the score for CTA, we compare the errors that \textit{gpt-4o} makes when we use zero-shot prompting compared to using the refined definitions in the prompt. We report the labels with more than 5 errors in Table~\ref{tab:errors-ref} for the SOTAB V2 dataset. In zero-shot prompting, we have 10 labels that have more than 5 errors, while this number decreases to 7 when using the refined definitions. Overall the inclusion of the refined definitions helps in decreasing the errors in 8 out of 10 high-errors labels. 
These decreases in errors contribute to the increase of 4.5\% F1 score compared to zero-shot prompting. Regarding the WikiTURL dataset, we notice that we have a high number of False Positives (FPs) errors. The CTA problem in WikiTURL is multi-label, and the model can choose to predict more than one label per column.
Comparing zero-shot prompting and refined definitions, we notice that when we include the definitions into the prompt, the model is more reluctant to predict more labels, making the precision of the model higher and bringing the large increase of 4.9\% in the Hamming Score compared to the zero-shot scenario (see Table~\ref{tab:definitions-turl}). Out of 5 labels that have the highest errors \textit{broadcast.artist}, \textit{broadcast.broadcast}, \textit{people.person}, \textit{music.musical\_group}, \textit{organization.organization} and \textit{film.film\_director}, 4 out of 5 labels are improved by at least 8 errors when using the \textit{refined} definitions.

\begin{table}
  \caption{Number of generation and inference tokens and their costs, and annotation cost per column for knowledge prompting with demonstration, refined and comparative definitions on gpt-4o.}
  \label{tab:cost-defs}
  \begin{tabular}{lrrrr}
    \toprule
     \textbf{Dataset} & & \textbf{demos} & \textbf{refined} & \textbf{comp.}\\
    \midrule
    \multirow{2}{*}{\textbf{SOTAB V2}} & Generation & 30K & 1,399K & 440K \\
    & Inference & 1,384K & 1,607K & 3,337K \\
    & Gener. Cost & \$0.07 & \$3.50 & \$1.10 \\
    & Infer. Cost & \$3.72 & \$4.27 & \$8.61 \\
    & Cost/Column & \$0.006 & \$0.007 & \$0.014 \\
    & F1 & 82.6 & \textbf{85.4} & 83.7 \\
    \midrule
    \multirow{2}{*}{\textbf{WikiTURL}} & Generation & 17K & 3,347K & 645K \\
    & Inference & 2,932K & 4,801K & 5,622K \\
    & Gener. Cost & \$0.04 & \$8.36 & \$1.61 \\
    & Infer. Cost & \$7.43 & \$12.1 & \$14.1\\
    & Cost/Column & \$0.010 & \$0.016 & \$0.019\\
    & F1 & 72.2 & \textbf{72.4} & 71.6 \\
    \midrule
    \multirow{2}{*}{\textbf{Limaye}} & Generation & 4K & 213K & 39K \\
    & Inference & 212K & 258K & 267K \\
    & Gener. Cost & \$0.01 & \$0.53 & \$0.10 \\
    & Infer. Cost & \$0.54 & \$0.65 & \$0.68\\
    & Cost/Column & \$0.005 & \$0.006 & \$0.006\\
    & F1 & 86.8 & \textbf{88.4} & 85.0 \\
  \bottomrule
\end{tabular}
\end{table}

\textbf{Costs.} We report the costs of generating and using \textit{demonstration}, \textit{refined} and \textit{comparative} definitions for the three datasets in Table~\ref{tab:cost-defs}. We report token usage separated into generation and inference tokens, where the former includes the tokens used to generate the definitions as well as the classification of the validation set, 
and the latter includes the tokens needed for classifying our test sets. Alongside the number of tokens, we report their respective total costs when using the \textit{gpt-4o-2024-08-06} model priced at \$2.5/1M tokens (as of January 2025). Lastly, we report the average inference cost per annotated column (Cost/Column in the table).
From the table, we can observe that \textit{demonstration} definitions have the lowest generation cost as well as the lowest inference cost. 
The difference between the inference costs of \textit{demonstration} and \textit{refined} definitions are similar for the three datasets (0.001-0.006 difference in cost per annotated column). This makes the \textit{refined} definitions the most performant method as the F1 score is also highest for this method on two out of three datasets.
On the other hand, \textit{comparative} definitions have a high inference cost compared to the other definitions, making this method not useful in applications with large number of tables.

\begin{figure}
  \centering
  \includegraphics[width=\linewidth]{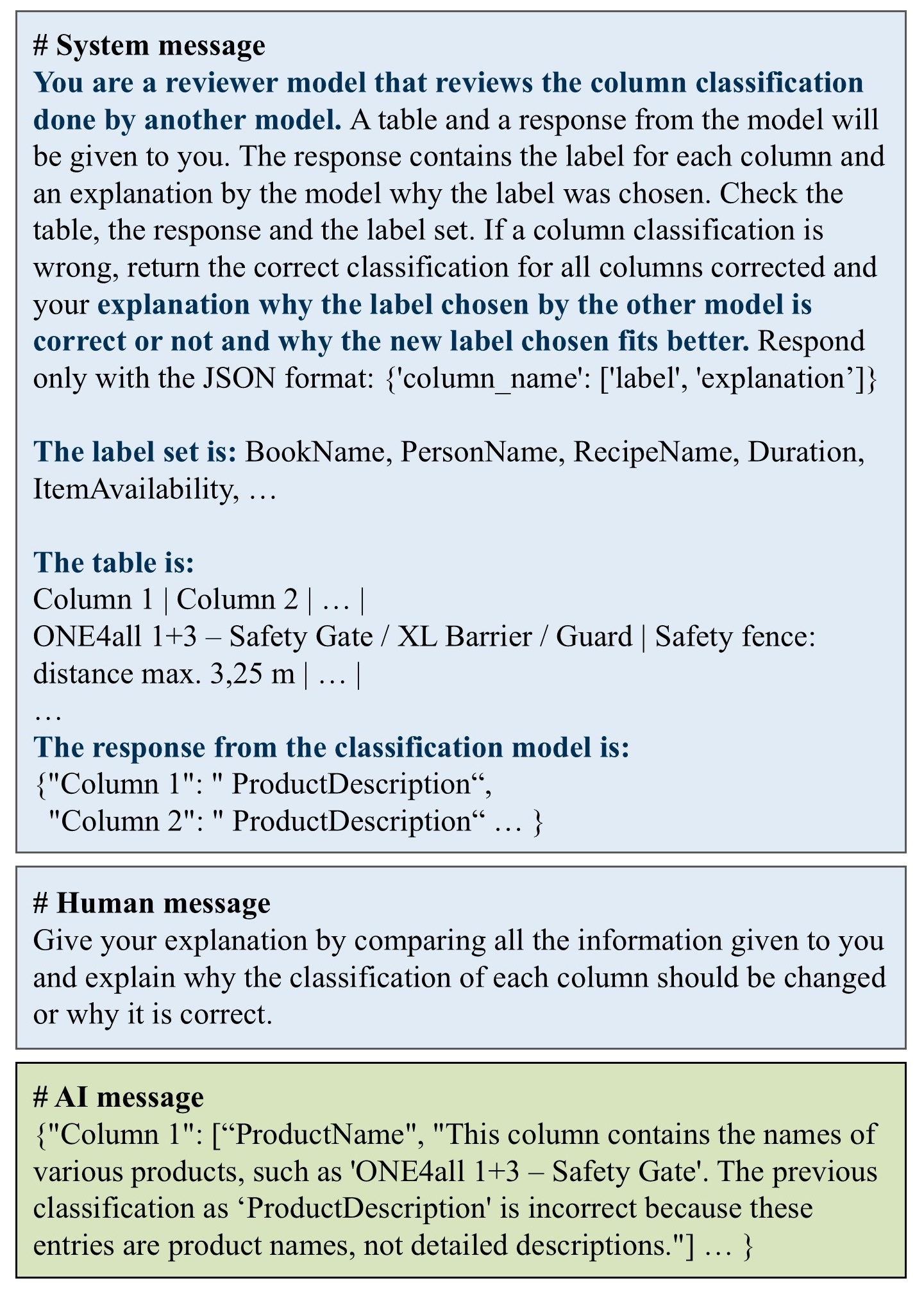}
  \caption{Self-correction: Example of a reviewer model prompt and its response (AI Message).}
  \label{fig:self-corr}
\end{figure}

\section{Self-Correction}
\label{sec:correction}
In this section, we test a \textit{self-correction}~\cite{pan2023automatically} strategy for correcting an LLM's initial classification. For this approach, we apply a two-step pipeline. 
In the first step, we instruct the model to annotate the columns of the test set. In the second step, we initialize a model with the role of a reviewer, and we feed to this reviewer model the label space, test tables and the classification done by the previous model for the tables. This is done one table at a time, meaning the we pass in one prompt only one input table. 
We prompt this model to generate a review where it should explain for each test column the reason why the previous model is correct or incorrect on choosing a certain label. 
The JSON format that we expect the model to generate the answer is \{column\_name: [label, explanation]\}, where explanation contains the review of the model for each column. The prompt in the first step follows the same zero-shot prompt of Figure~\ref{fig:zero-shot}. An example of a prompt for the reviewer model in the second step is shown in Figure~\ref{fig:self-corr}. In the \textit{system message}, we pass the task description as well as some task instructions that detail in which format the model should answer. We then pass the label set, input table (from the test set) and the response from the classification model in step one. In the \textit{human message} that follows we ask the model to give a review based on the information given in the previous message.

In our experiments, we test the reviewer model in three scenarios: (i) zero-shot scenario: shown in Figure~\ref{fig:self-corr}, we show the model the label set, the input table and the classification of zero-shot prompting (from Section~\ref{sec:baselines}) for this input table. (ii) knowledge prompting scenario with \textit{demonstration} definitions: in addition to the previous information in the zero-shot scenario, we provide to the reviewer model the previously generated \textit{demonstration} definitions and the classification of the model using these definitions from Section~\ref{sec:labels}. (iii) knowledge prompting scenario with selected \textit{comparative} definitions: similar to the last scenario, we provide \textit{comparative} definitions in the prompt. However we do not provide the whole set of comparative definitions. We provide selected definitions based on the first model's classification response.  For example, in Figure~\ref{fig:self-corr} the classification model of the first step has classified the columns in this table with the label ``ProductDescription". Based on this classifiation we provide the comparative definition of the label ``ProductDescription" i.e. tips how to distinguish the ``ProductDescription" label from other labels. The idea behind selecting these definitions is to make the model reason on the different labels that the first model could have mistaken the ``ProductDescription" label with. In this case the classification responses included in the prompt are taken from the results of zero-shot prompting in Section~\ref{sec:baselines}.

\begin{table}
  \caption{Micro-F1 results of the self-correction pipeline for SOTAB and Limaye and Hamming Score for WikiTURL.}
  \label{tab:critic}
  \begin{tabular}{llcccc}
    \toprule
    \textbf{Dataset} & \textbf{Prompt} & \textbf{Ll-8B} & \textbf{Ll-70B} & \textbf{gpt-mini} & \textbf{gpt-4o}  \\
    \midrule
     \multirow{3}{*}{SOTABV2} & 0-shot & 53.6 & 68.6 & 68.5 & 81.6 \\
     & demos. & \textbf{58.4} & \textbf{68.9} & \textbf{71.8} & 84.6 \\
     & compar. & 53.0 & 68.5 & 69.2 & \textbf{87.1 }\\\midrule
     & $\Delta$0-shot & -2.4 & +1.2 & -0.9 & +1\\
     & $\Delta$demos. & +0.1 & -1.5 & -1.1 & +2\\
     & $\Delta$comp. & -2.3 & +0.9 & -1.1 & +3.4 \\
     \midrule
     \multirow{3}{*}{Limaye} & 0-shot & 55.9 & 73.4 & 77.8 & 82.9 \\
    & demos. & 56.5 & 67.3 & 74.5 & \textbf{88.5} \\
    & compar. & \textbf{60.8} & \textbf{81.3} & \textbf{81.7} & 84.9 \\\midrule
    & $\Delta$0-shot & -10 & -3.3 & +1.3 & +0.3 \\
     & $\Delta$demos. & -9.6 & -14.3 & -3.2 & +1.7 \\
     & $\Delta$comp. & -16.5 & +4.7 & +1.2 & -0.1 \\
     \midrule
    \multirow{3}{*}{WikiTurl} & 0-shot & 24.3 & \textbf{51.7} & 51.7 & 63.0 \\
     & demos. & \textbf{30.9} & 47.9 & 54.1 & \textbf{66.8} \\
     & compar. & 23.5 & 51.4 & \textbf{55.3} & 63.2 \\\midrule
     & $\Delta$0-shot & -1.2 & -0.8 & +1.6 & -0.2 \\
     & $\Delta$demos. & +2.1 & +0.2 & +1.9 & -1.1 \\
     & $\Delta$comp. & -2.1 & +6.5 & -1.8 & -5.5 \\
  \bottomrule
\end{tabular}
\end{table}

\textbf{Results and Costs.} The results of the three setups of the self-correction pipeline are reported in Table~\ref{tab:critic}. In the table, we report the difference to each method's one-step counterpart, for example we report the difference between 0-shot prompting to 0-shot self-correction ($\Delta$0-shot). Generally, the self-correction pipeline does not improve one-step CTA when tested on the Llama and gpt-mini models. On the larger \textit{gpt-4o} model we observe in Limaye and in SOTAB an increase of at least 1.7\% F1 when self-correction is paired with \textit{demonstration} definitions. We list the costs of the self-correction pipeline using the \textit{gpt-4o model} in Table~\ref{tab:cost-critic}. From the three setups, we notice that using selective \textit{comparative} definitions is more efficient token-wise than the \textit{demonstration} definitions as the cost per column (Cost/Column) is lower. Comparing self-consistency (see Table~\ref{tab:cost-zero-few-self}) and 0-shot self-correction, we see that the cost per column of self-correction is higher as the model gives an explanation of the changes it makes to the previous classification. The F1 score between these two methods is similar, however both methods are less efficient than one-step 0-shot prompting.

\begin{table}
  \caption{Number of generation and inference tokens and their costs, and annotation cost per column for 3 setups of the self-correction pipeline for gpt-4o.}
  \label{tab:cost-critic}
  \begin{tabular}{lrrrr}
    \toprule
     \textbf{Dataset} & & \textbf{0-shot} & \textbf{demos} & \textbf{comp.}\\
    \midrule
    \multirow{2}{*}{\textbf{SOTAB V2}} & Generation & - & 30K & 440K \\
    & Inference & 608K & 2863K & 846K \\
    & Gener. Cost & - & \$0.07  & \$1.10 \\
    & Infer. Cost & \$2.06 & \$7.93  & \$2.76 \\
    & Cost/Column & \$0.003 & \$0.012  & \$0.004 \\
    \midrule
    \multirow{2}{*}{\textbf{WikiTURL}} & Generation & - & 17K & 645K \\
    & Inference & 983K & 5897K & 1456K\\
    & Gener. Cost & - & \$0.04 & \$1.61 \\
    & Infer. Cost & \$2.87 & \$15.1 & \$4.09\\
    & Cost/Column & \$0.004 & \$0.02 & \$0.005\\
    \midrule
    \multirow{2}{*}{\textbf{Limaye}} & Generation & - & 4K & 39K \\
    & Inference & 64K & 430K & 83K \\
    & Gener. Cost & - & \$0.01  & \$0.10 \\
    & Infer. Cost & \$0.18 & \$1.09 & \$0.23\\
    & Cost/Column & \$0.001 & \$0.009 & \$0.001\\
  \bottomrule
\end{tabular}
\end{table}

\section{Fine-tuning}
\label{sec:ft}

In this section, we experiment with fine-tuning the four LLMs tested in previous sections and investigate whether it is beneficial to use the previously generated definitions within the fine-tuning process. 
The goal of integrating the definitions is to help the model adapt to the specific meanings that the labels have in a dataset. 
We compare this setup to the simple fine-tuning scenario, where we fine-tune the LLMs only on the CTA task and check if this simple fine-tuning is enough to adapt the model to the datasets' vocabularies.

We build three different training sets for fine-tuning: 

\begin{figure}
  \centering
  \includegraphics[width=\linewidth]{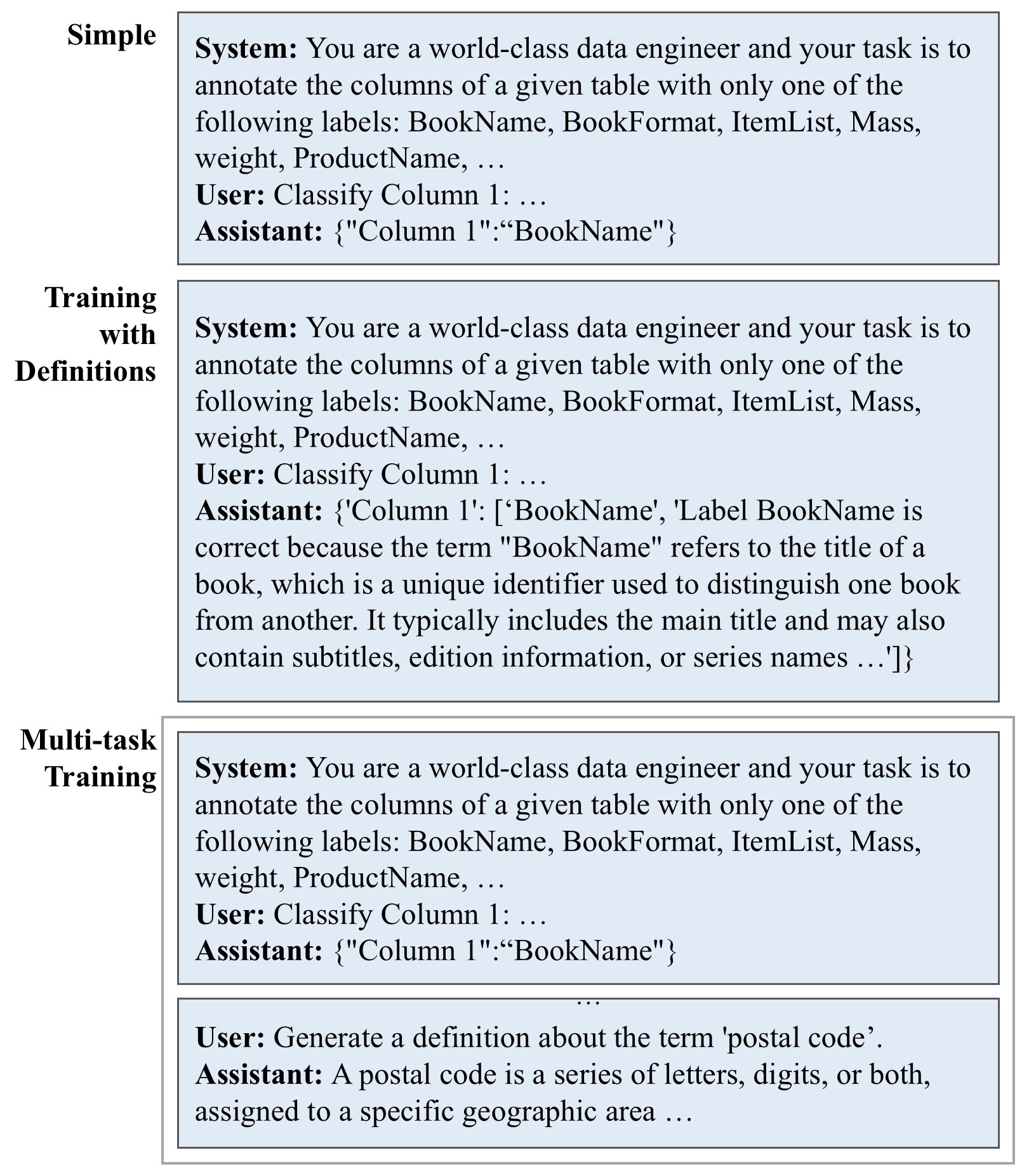}
  \caption{Training examples from the three fine-tuning sets.}
  \label{fig:ft-sets}
\end{figure}

\begin{table*}
  \caption{Fine-tuning results on SOTAB V2 and WikiTURL using zero and knowledge prompting. Numbers in bold represent the method with the highest score for each prompting technique. We report the differences of the best zero-shot prompting result of fine-tuned models to the best prompting with definitions results of fine-tuned models ($\Delta$ 0-shot fine-tuned).}
  \label{tab:ft-res}
  \begin{tabular}{l|cccc|cccccccc}
    \toprule
     \textbf{\multirow{5}{*}{Setup}} & \multicolumn{4}{c|}{\textbf{SOTAB V2}} & \multicolumn{8}{c}{\textbf{WikiTURL}} \\\midrule
      & \textbf{Ll-8B} & \textbf{Ll-70B} & \textbf{gpt-mini} & \textbf{gpt-4o} & \multicolumn{2}{c}{\textbf{Ll-8B}} & \multicolumn{2}{c}{\textbf{Ll-70B}} & \multicolumn{2}{c}{\textbf{gpt-mini}} & \multicolumn{2}{c}{\textbf{gpt-4o}}  \\
     & F1 & F1 & F1 & F1 & F1 & HS & F1 & HS & F1 & HS & F1 & HS \\
    \midrule
     \textbf{Zero-shot Prompt} & & & & & & & & & & & &\\
    \midrule
     simple fine-tuning & 77.9 & 86.4 & 87.0 & 87.8 & 62.5 & 57.8 & 64.3 & 61.1 & \textbf{67.3} & \textbf{64.1} & 71.1 & 67.4\\
     fine-tuning-with-definitions & 71.4 & 85.4 &  - & - & 60.0 & 57.4 & 64.0 & 61.1 & - & - & - & - \\
     multi-task fine-tuning & 77.3 & \textbf{87.8} &  87.0 & - & \textbf{64.1} & \textbf{60.0} & \textbf{66.4} & \textbf{63.0} & 64.4 & 61.2 & - & - \\
     multi-task-with-demonstrations & \textbf{80.7} & 86.9 & - & - & 63.1 & 59.2 & 60.2 & 57.3 & - & - & - & - \\
     \midrule
     $\Delta$ 0-shot (not fine-tuned) & +24.7 & +20.4 & +17.6 & +6.9 & +32.3 & +34.5 & +9.1 & +10.5 & +6.3 & +12.9 & +1.1 & +4.2 \\
     \midrule
     \textbf{Prompts with Definitions} & & & & & & & & & & & &\\
    \midrule
     fine-tuned + comparative defs & 80.8 & 85.8 & \textbf{89.1} & 90.0 & \textbf{70.1} & \textbf{66.1} & \textbf{65.6} & \textbf{62.7} & 65.4 & 62.0 & \textbf{74.1} & \textbf{71.0} \\
     fine-tuned + demonstration defs & 80.3 & 85.6 & 87.0 & 89.1 & 65.0 & 60.8 & 65.0 & 62.1 & 66.4 & 63.1 & 72.3& 68.7 \\
     fine-tuned + refined defs & \textbf{81.3} & \textbf{86.1} & 87.2 & \textbf{91.8} & 65.1 & 61.0 & 64.4 & 61.4 & 63.3 & 60.2 & \underline{74.0} & \underline{70.3} \\\midrule
     $\Delta$ 0-shot fine-tuned & +0.6 & -1.7 & +2.1 & +4.0 & +6.0 & +6.1 & -0.8 & -0.3 & -0.9 & -1 & +3.0 & +3.6 \\
  \bottomrule
\end{tabular}
\end{table*}

\begin{enumerate}
    \item\textbf{Simple fine-tuning.} In the simple fine-tuning set, we fine-tune the models on the CTA task using our zero-shot prompts shown in Figure \ref{fig:zero-shot}. Based on initial experiments, we leave out the instructions when we fine-tune the models as they do not impact the results of the model and keep the instructions only when fine-tuning the OpenAI models for SOTAB V2. For this set, we expect the model to learn to generate the answers in the following JSON format: \{column\_name: label\}.
    \item\textbf{Fine-tuning with definitions.} With the motivation that including the label definitions into the training process can help the model adjust better to the dataset vocabulary, we modify the simple fine-tuning set to include an explanation in the JSON response alongside the correct label for the column. The explanations are constructed by using the label definitions as the reason why a label is correct for annotating a column. We build the explanation by combining a prefix with the label definition: ``Label \textit{[label]} is correct because the term \textit{[label definition]}".
    The label definitions that we use are the \textit{demonstration definitions} that we generated in Section~\ref{sec:labels}. An example can be seen in the second part of Figure~\ref{fig:ft-sets}. In the \textit{assistant message} that the model should learn to generate, \textit{Column 1} is classified as BookName and the explanation why this label is the correct annotation is the definition of BookName. The JSON response that the model should learn to generate in this setup is \{column\_name: [label, explanation]\}. 

    \item\textbf{Multi-task fine-tuning.} In this set, 
    we keep the two tasks of CTA and definition generation separated. We combine the \textit{simple} set with a \textit{definition generation} set, 
    where the model should learn 
    to generate the \textit{demonstration} definitions. For example, for the SOTAB V2 dataset, this fine-tuning set has a total of 748 examples, 698 of which are the annotated tables of the CTA training set (shown in Table~\ref{tab:datasets}), while 50 are the definition generations which corresponds to the total number of labels in the SOTAB V2 dataset.
    In Figure \ref{fig:ft-sets}, in the last block we show two examples from the set, one which shows an example of the CTA task where the model should classify the columns of a table, and the other shows the definition generation task where the model should generate the definition of the term ``postal code". 
    In the figure, we show for the definition generation task a prompt without demonstrations. In our experiments, we also test a version of the multi-task training set where we include some demonstrations in the definition generation prompt. We name this variation as \textit{multi-task with demonstrations}.
    For the CTA task, the model should generate the answer in the JSON format \{column\_name: label\}, while definitions generation is a text generation task.
\end{enumerate}

To test the fine-tuned models, we use two prompting techniques: \textbf{zero-shot} and \textbf{knowledge generation} prompting. In zero-shot prompting, we use the zero-shot prompt in Figure \ref{fig:zero-shot}. In knowledge generation prompting, we use the methods described in Section~\ref{sebsec:def-gen} to generate \textit{demonstration}, \textit{comparative} and \textit{refined} definitions. 
For each dataset, we use the corresponding \textit{gpt-4o} model fine-tuned on the best performing set to generate the definitions. 
By testing knowledge prompting, the goal would be to investigate whether augmenting prompts with generated definitions can help further adapt the fine-tuned models to the dataset vocabularies, or if the knowledge is already gained through the fine-tuning process. 

\textbf{Fine-tuning Details.} When fine-tuning open-source models, we use a learning rate of 1e-4, a batch size of 16 for the small Llama-8B model and a batch size of 8 for the larger Llama-70B model. We train the models for 10 epochs using a maximum input length of 5020. We use QLoRA~\cite{dettmers2024qlora} for fine-tuning and test three LoRA ranks and alphas [8, 16, 32] as well as LoRA dropouts [0, 0.05, 0.1]. From our hyperparameter search, we notice that a rank and alpha of 32 and a dropout of 0.1 achieves the highest performance for both datasets and both Llama models. The OpenAI models are trained using the default parameters for 3 epochs and a batch size of 1.

\textbf{Results.} The results of the fine-tuning experiments are reported in Table \ref{tab:ft-res}. In the first part of the table we present the results of fine-tuning LLMs using our four fine-tuning sets and testing them using zero-shot prompting. Fine-tuning all models brings a large increase in the metrics for both datasets, from 6.3\% up to 32.3\% in the F1 score. The largest increases (>20\%) can be noticed for the Llama-8B model on both datasets. There is one exception for the WikiTURL dataset and the \textit{gpt-4o} model, where the increase from the zero-shot scenario is only 1.1\%. This small increase can be attributed to the multi-label problem, where in the training set we have different combinations of labels than in the test set. 

Fine-tuning the Llama models with the \textit{multi-task} sets, there is a small increase in F1 score compared to fine-tuning with the \textit{simple} sets, on average a 2\% F1 increase on both datasets.
On the other hand, using the definitions as explanations in the \textit{fine-tuning-with-definitions} set hurts the model performance when compared to the \textit{simple} set. This set is also the most expensive token-wise, as the label definitions are repeated as many times as the number of examples in the training set. For the OpenAI models, due to limited resources for fine-tuning the models, we test on \textit{gpt-mini} only the \textit{multi-task} set that brought a small increase to the Llama models. We notice from these runs that this set does not improve but can also harm the performance of the model, as seen when fine-tuning \textit{gpt-mini} with the WikiTURL dataset. We therefore only run simple fine-tuning with the large \textit{gpt-4o} model.

In the second part of Table~\ref{tab:ft-res}, we report the results of using knowledge prompting on the highest performing fine-tuned model for each model and dataset. 
From the results, we notice that we have a positive impact on the large \textit{gpt-4o} model: for both datasets we have an increase of at least 3\% in F1 when we use the error-based definitions: \textit{comparative} and \textit{refined} definitions. Using these definitions on the smaller fine-tuned models, does not impact the performance of the models, and we conclude that the models have already gained some knowledge about the vocabulary from fine-tuning and the additional definitions do not impact the annotation. There is one exception for the small fine-tuned 8B model on the WikiTURL dataset, when paired with \textit{comparative} definitions the model reaches the same performance as fine-tuning the large \textit{gpt-4o} model with the simple fine-tuning set.

\begin{table}
    \centering
    \caption{SOTAB V2 labels with more than 5 errors for the fine-tuned models tested with 0-shot and knowledge prompting with refined definitions on gpt-4o.}
    \begin{tabular}{lcccc}
    \toprule
    \textbf{Label} & \textbf{ft-0-shot} & \textbf{ft-refined} & \textbf{$\Delta$ft-0-shot} \\\midrule
        Duration & 21 & 0 & -21  \\
         weight  & 13 & 11 & -2 \\
         Identifier  & 11 & 8 & -4 \\
         QuantitativeValue  & 10 & 10 & 0\\
         unitCode  & 9 & 8 & -1\\
         \bottomrule
    \end{tabular}
    \label{tab:errors-ft}
\end{table}

\textbf{Error Analysis.} In Table~\ref{tab:errors-ft}, we report for the fine-tuning scenario the labels with more than 5 errors of gpt-4o for \textit{zero-shot} and knowledge prompting with \textit{refined} definitions on the SOTAB V2 dataset. The number of errors is less than in the previous non-fine-tuning scenarios, amounting to 5 labels. From the table we can observe that using the \textit{refined} definitions, we have decreases in 4 out of 5 labels, with labels such as weight, QuantitativeValue and unitCode still having more than 5 errors and recurring from previous sections. We also notice an improvement of the long tail labels: in the zero-shot scenario we have 21 labels with no errors, and this number improves to 30 labels when \textit{refined} definitions are used. Overall, these improvements in the number of errors brings the 4\% increase to the SOTAB V2 dataset, bringing the F1 score to 91.8\%. For WikiTURL, we observe similar patterns as in the non-fined-tuning scenario. We have less false positives when we use the \textit{refined} definitions as the model does not predict as many labels per column as with zero-shot prompting, which brings an increase of 2.9\% in Hamming Score and in the F1 score. Out of 6 labels with most errors, all of them improve when using \textit{refined} definitions.

\begin{table}
  \caption{SOTAB V2 and WikiTURL token usage analysis for the gpt-4o model for fine-tuning scenario tested with zero-shot prompting and with error-based definitions.}
  \label{tab:cost}
  \begin{tabular}{lrrrc}
    \toprule
     \textbf{Dataset} & & \textbf{0-shot} & \textbf{refined} & \textbf{comp.}\\
    \midrule
    \multirow{2}{*}{\textbf{SOTAB V2}} & FT & 1,895K & 1,895K & 1,895K \\ 
    & Generation & - & 928K & 385K \\
    & Inference & 268K & 1,023K & 1,383K \\
    & FT Cost & \$47.4 & \$47.4 & \$47.4 \\
    & Gener. Cost & - & \$3.48 & \$1.45 \\
    & Infer. Cost & \$1.28 & \$3.90 & \$5.26 \\
    & Cost/Column & \$0.002 & \$0.007 & \$0.009 \\
    & F1 & 87.8 & \textbf{91.8} & 90.0 \\
    \midrule
    \multirow{2}{*}{\textbf{WikiTURL}} & FT & 801K & 801K & 801K \\ 
     & Generation & - & 2,912K & 626K \\
    & Inference & 473K & 3,799K & 4,457K \\
    & FT Cost & \$20.0 & \$20.0 & \$20.0 \\
    & Gener. Cost & - & \$10.9 & \$2.35 \\
    & Infer. Cost & \$1.88 & \$14.3 & \$16.8 \\
    & Cost/Column & \$0.002 & \$0.020 & \$0.023 \\
    & HS & 67.4 & 70.3 & \textbf{71.0} \\
  \bottomrule
\end{tabular}
\end{table}

\textbf{Costs.} In Table~\ref{tab:cost}, we report the token usage and costs for testing the fine-tuned \textit{gpt-4o} models in a zero-shot scenario as well as in a knowledge prompting scenario with \textit{refined} and \textit{comparative} definitions. We report the generation, inference and fine-tuning tokens. As before, the generation tokens include the tokens used for generating/refining the definitions as well as classifying the validation set. The inference tokens include the tokens used to classify the test set, while the fine-tuning tokens include the total tokens used in fine-tuning. The cost of fine-tuning the model is \$25/1M training tokens and the usage of the fine-tuned gpt-4o model is \$3.75/1M tokens (as of January 2025). The Cost/Column denotes the inference cost per annotation.

\textit{Fine-tuning or knowledge prompting without fine-tuning?} Comparing the previous methods from Table~\ref{tab:cost-defs} to the 0-shot fine-tuning scenario, we can see that using knowledge prompting with \textit{refined} definitions in a non-fine-tuning scenario has a lower total cost for SOTAB V2 and WikiTURL than the fine-tuning scenario. However, the cost per column of the 0-shot fine-tuning scenario which is 0.002 for both datasets is lower than the cost per column of the \textit{refined} definitions which stands at 0.007 and 0.016 for SOTAB V2 and WikiTURL respectively (from Table~\ref{tab:cost-defs}). We can conclude from this that in scenarios with small datasets such as SOTAB V2 and WikiTURL, knowledge prompting is more efficient cost-wise and performance-wise reaches a similar score to fine-tuning. In scenarios with a large amount of tables and columns, 0-shot fine-tuning becomes more efficient. For example, if the amount of columns in a dataset exceeds 9400\footnote{Calculated using the Cost/Column of both methods and fine-tuning costs of the SOTAB V2 dataset}, the fine-tuning scenario becomes the cheaper of the two options.

\textit{Zero-shot or Knowledge Prompting when using fine-tuned models?} The highest F1 scores in the fine-tuning scenario for both SOTAB V2 and WikiTURL are reached when fine-tuned models are tested using error-based definitions-augmented prompts. The sets of \textit{comparative} and \textit{refined} definitions have a small difference in F1 score between them, however from the inference cost we can conclude that \textit{refined} definitions are the most efficient token-wise and performance-wise between them. The cost per annotated column is also higher for \textit{comparative} definitions, which means that in applications with large amount of tables, this method is more expensive.

\section{Conclusion}
\label{sec:conclusion}
In this paper, we aim to evaluate knowledge generation and self-refining strategies for LLM-based CTA. The methods that we test include generation of label definitions with knowledge prompting, self-correction and self-refining generated label definitions. We conclude that there is no best strategy and the best performing method depends on the dataset/model combination. We show that generated definitions increase the F1 score of the models in 11 out of 12 cases by an average of 2.4\% compared to zero-shot prompting, while using our error-based self-refinement method to enhance definitions brings an additional average increase of 3.9\% in 10 out of 12 cases. In our fine-tuning setup, we test integrating the label definitions into the fine-tuning process and find that this integration can bring an average increase of 2\% to the Llama models. We further conclude that fine-tuning the LLMs is more efficient 
for use cases requiring a large amount of tables to be annotated (>9000 columns) compared to using knowledge prompting with refined definitions in the non-fine-tuning scenario. The ability of the LLMs for self-correction proved useful for the large gpt-4o model when paired with error-based definitions. For the other models and for zero-shot prompting the results were not consistent and the additional self-correction step mostly harmed the performance of the model. Finally, using the fine-tuned models in combination with the self-refinement method, increased the F1 score of two datasets on the \textit{gpt-4o} model by at least 3\% compared to zero-shot prompting the fine-tuned model. 

\section{Artifacts}
\label{sec:artifacts}
We make our code, prompts, data, generated definitions and responses from the LLMs available in our Github repository\footnote{\url{https://github.com/wbsg-uni-mannheim/TabAnnGPT}} in the folder ``KnowledgeSelfRefinementForCTA".

\bibliographystyle{ACM-Reference-Format}
\bibliography{edbt-2026}

\end{document}